\documentclass[10pt,twocolumn,letterpaper]{article}

\usepackage{cvpr}
\usepackage{times}
\usepackage{epsfig}
\usepackage{graphicx}
\usepackage{amsmath}
\usepackage{amssymb}

\usepackage{subfig}
\usepackage{xcolor}

\usepackage[pagebackref=true,breaklinks=true,letterpaper=true,colorlinks,bookmarks=false]{hyperref}

\cvprfinalcopy 

\def\cvprPaperID{7919} 

\ifcvprfinal\fi
\begin{document}

\makeatletter
\let\@oldmaketitle\@maketitle
\renewcommand{\@maketitle}{\@oldmaketitle
  \includegraphics[width=\linewidth]
    {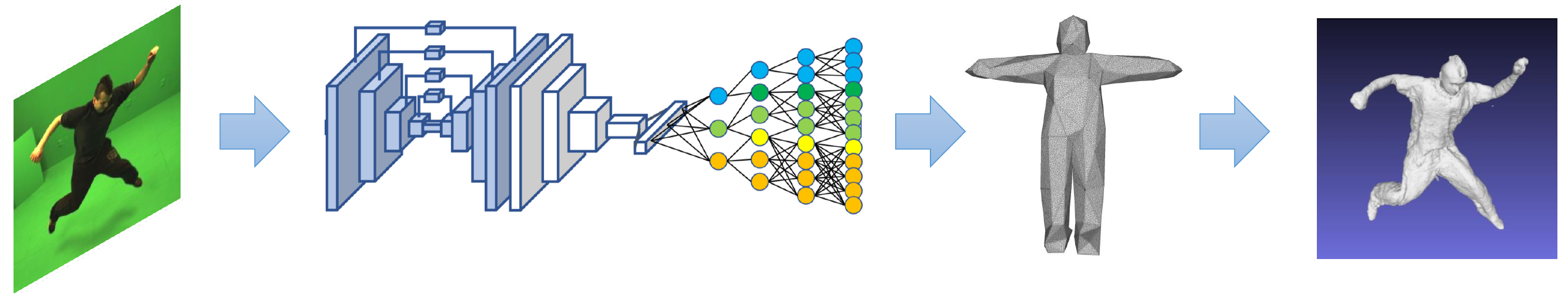}
    \captionof{figure}{We present a tetrahedral volumetric representation of the human body and a method called TetraTSDF that is able to retrieve the detailed 3D shape of a person wearing loose clothes from a single 2D image.
    }
     \rule{\textwidth}{0.4pt}
    \label{fig:front}
    }
\makeatother

\title{TetraTSDF: 3D human reconstruction from a single image with a tetrahedral outer shell}

\author{Hayato Onizuka \\
Kyushu University\\
{\tt\small mound028810@gmail.com}
\and
Zehra Hay{\i}rc{\i} \\
Technical University of Munich\\
{\tt\small zehrahayirci@gmail.com}
\and
Diego Thomas \\
Kyushu University\\
{\tt\small thomas@ait.kyushu-u.ac.jp}
\and
Akihiro Sugimoto \\
National Institute of Informatics\\
{\tt\small sugimoto@nii.ac.jp}
\and
Hideaki Uchiyama \\
Kyushu University\\
{\tt\small uchiyama@limu.ait.kyushu-u.ac.jp}
\and
Rin-ichiro Taniguchi \\
Kyushu University\\
{\tt\small rin@kyudai.jp}
}

\maketitle
\thispagestyle{empty}

\begin{abstract} \vspace{-5.2pt}
   Recovering the 3D shape of a person from its 2D appearance is ill-posed due to ambiguities. Nevertheless, with the help of convolutional neural networks (CNN) and prior knowledge on the 3D human body, it is possible to overcome such ambiguities to recover detailed 3D shapes of human bodies from single images. Current solutions, however, fail to reconstruct all the details of a person wearing loose clothes. This is because of either (a) huge memory requirement that cannot be maintained even on modern GPUs or (b) the compact 3D representation that cannot encode all the details. In this paper, we propose the tetrahedral outer shell volumetric truncated signed distance function (TetraTSDF) model for the human body, and its corresponding part connection network (PCN) for 3D human body shape regression. Our proposed model is compact, dense, accurate, and yet well suited for CNN-based regression task. Our proposed PCN allows us to learn the distribution of the TSDF in the tetrahedral volume from a single image in an end-to-end manner. Results show that our proposed method allows to reconstruct detailed shapes of humans wearing loose clothes from single RGB images.
\end{abstract}

\section{Introduction}
\label{sec:intro}
Detailed 3D shapes of the human body reveal personal characteristics that cannot be captured with standard 2D pictures. Such information is crucial for many applications in the entertainment industry (3D video), business (virtual try-on) or medical use (self-awareness or rehabilitation). The first systems that built 3D models of the human body were designed to work in controlled settings using laser scanners, multi-view calibrated camera arrays or markers. These systems were hard to set up, not affordable, and offering limited application areas.

In the last decade, consumer-grade depth cameras have been successfully used to build 3D models of the human body \cite{yu2018doublefusion}. However, high quality depth cameras are still not available to a large part of the consumers (most of smartphones are still not equipped with depth cameras). Moreover, consumer-grade depth cameras do not work well in outdoor environments. As a consequence, methods that can efficiently reconstruct detailed 3D human shapes in unconstrained environments are needed. In this work, we focus on the task of detailed 3D human body reconstruction in a single shot with a standard RGB camera.

In the literature, there are two strategies to generate a 3D model from a single color image: (1) parametric model fitting and (2) 3D model regression. Methods that fall into the first strategy fit a parametric human template 3D model (such as the SMPL model \cite{SMPL:2015}) to the input color image. To fit the template model, various cost functions have been proposed that consider silhouette, skeleton and feature points (\cite{Bogo:ECCV:2016}). The latter strategy takes advantage of the recent advances in convolutional neural networks (CNN). Thereby, depth image regression is followed by volumetric fusion \cite{tatarchenko2016multi}, or end-to-end RGB to 3D model techniques \cite{Alldieck_2019_ICCV, Saito_2019_ICCV, varol18_bodynet} have been proposed.

CNN-based methods are promising for reconstructing 3D human bodies from a single color image because they have the potential to capture detailed and complex features (such as clothes wrinkles). However, several limitations exist. The main problem stems from the volumetric TSDF representation which is used to regress the 3D shape. The resolution of the volumetric representation has to meet the fine details of the human body, meaning that a considerable amount of memory is required. However, such memory constraint is hard to be maintained even on modern GPUs.

The key challenge for achieving higher accuracy in 3D human body shape reconstruction is to define a more compact 3D human body representation in memory that still allows casting the problem as a well-adapted regression task. One solution may be displacement mapping \cite{Alldieck_2019_ICCV}. However, this compact representation inevitably loses some details in shape because its dependency to the SMPL mesh. For instance, occluded parts in non-convex areas or garments like shoes or gloves cannot be reconstructed by using displacement mapping.

In this paper, we propose a new volumetric 3D body representation for end-to-end 3D body shape regression from a single color image. Our proposed 3D body representation is based on a tetrahedral TSDF field embedded into a human-specific outer shell. The outer-shell is built from a coarse version of the SMPL model \cite{SMPL:2015} and can be fitted to a human body using the SMPL pose and shape parameters. The tetrahedral TSDF field is built at the summits of a tetrahedral volumetric grid defined by the outer shell. We also propose a new network to estimate the tetrahedral TSDF field from a single color image that combines CNN and our proposed Part Connection Network (PCN).

Our contributions are three fold: (1) a new 3D body volumetric representation that is compact, dense, accurate and yet well suited for CNN-based regression tasks; (2) a method to generate high quality TSDF fields from ground truth (GT) 3D human body scans; and (3) a new CNN-PCN based hourglass network for end-to-end regression of 3D human body shape from a single color image.

\section{Related work}
\label{sec:relWorks}
Fitting a parametric 3D model to the input 2D color image has been the standard way to reconstruct a 3D shape from a single 2D image for a long time. Recently, using CNNs has proven to be a powerful alternative. Here, we review related works that use both of these strategies, with a particular focus on the human body shape reconstruction problem.

\subsection{Template model fitting}
The classic approach to estimate a 3D shape of an object from a single color image is to fit a template 3D model so that it matches its 2D projection while satisfying some constraints (\textit{e.g.}, \cite{ Bogo:ECCV:2016, cashman2012shape, SMPL:2015}). Landmark-guided non-rigid registration of 3D templates to 2D or 2.5D inputs have been widely studied. Lu \textit{et.al.} \cite{lu2008deformation}, for example proposed to use facial landmarks to fit a deformable face model to 2.5D data. Cashman \textit{et.al.} \cite{cashman2012shape}, proposed to represent a 3D deformable model with a linear combination of subdivision surfaces. Such model can be fitted to a collection of 2D images by manually providing some key-points and the silhouette of the object to be reconstructed.

Recovering the 3D human shape from a single RGB image is an open-problem in the field. There are only few proposed techniques that deal with complex poses and deformations. Most of the techniques rely either on a pre-scanned model of the subject \cite{de2008performance, habermann2019livecap, kraevoy2009modeling, tan2010image, vlasic2008articulated} or a template model \cite{bualan2008naked, balan2007detailed}. In \cite{habermann2019livecap}, the authors propose to first scan the 3D model of a person using a multi-view reconstruction system. Then, the reconstructed 3D model is non-rigidly aligned to the RGB video in real-time. Similarly, in \cite{guan2009estimating}, Bogo \textit{et.al.}~\cite{Bogo:ECCV:2016}, propose to optimize the parameters of a parametric template model given input images and poses by using many cues like silhouette overlap, height constraint and smooth shading. The authors use the SMPL model \cite{SMPL:2015} to recover various parameters such as pose and shape from a single RGB image. Recently, 
Kolotouros \textit{et al.}~\cite{kolotouros2019learning} proposed a method to estimate SMPL parameters by developing a self-improving loop combining CNN and optimization method. However, template-based methods fail to capture loose clothing and thus only reconstruct bare human body.

\subsection{Convolutional Neural Network regression}
Recently, CNNs have brought new possibilities to many domains in computer vision. 3D shape reconstruction from a single image is one of those areas that strongly developed with the availability of new CNN tools. 



Inspired by the extraordinary performance of CNNs for segmentation tasks, several methods have been proposed that represent the 3D shapes as binary occupancy maps \cite{choy20163d, girdhar2016learning, tulsiani2017multi, wu2017marrnet, wu2016learning}. If the task of estimating the 3D surface is expressed as a segmentation problem, then CNNs can predict outside and inside voxels. For example, Wu et al. \cite{wu20153d} proposed an extension of 2D CNNs for the case of volumetric outputs. Further optimization improvements were proposed in \cite{yan2016perspective} and \cite{zhu2017rethinking}. In the case of 3D face regression, Jackson \textit{et.al.}, \cite{Jackson2017LargeP3} proposed a method for direct regression of a volumetric representation of the face using CNN. By using probabilistic assignment, smooth surfaces could be obtained. 
Varol \textit{et.al.} \cite{varol18_bodynet}, extended this method to full body shape regression. All these methods share the common limitation that the memory consumption scales cubically with the shape resolution. Even with modern GPUs, volumetric regression networks only work with low resolution grid. Then, only coarse 3D models can be generated.

Riegler \textit{et al.}~\cite{riegler2017octnet}, proposed to use octrees to reduce the memory usage and adapt the CNNs to predict high-resolution occupancy maps providing the tree structure is known in advance. However, the method can not be applied to reconstruct 3D human bodies with different poses because the tree structure changes with every new input. In \cite{tatarchenko2017octree}, Tararchenko \textit{et.al.} proposed a technique to overcome this problem by also predicting the tree structure. However, training the network to learn the sparse structure of octrees is effortful. Recently, Saito \textit{et al.}~\cite{Saito_2019_ICCV} proposed a memory-efficient method by handling each 3D point individually. They reported high accuracy 3D reconstruction results on a private dataset while using reasonable amount of memory. However, this method jointly estimates the 3D shape and body pose while 3D body pose estimators have made significant progress recently and achieved high accuracy results (\eg, \cite{iskakov2019learnable} reported an average error of less than $2$ cm for the 3D joint position). We reason that the tasks should be separated: one task for body pose estimation and one task dedicated to the 3D shape estimation.

Recently, Alldieck \textit{et al.}~\cite{Alldieck_2019_ICCV} have proposed to use displacement mapping on top of a template human model to represent the 3D human body with loose clothes. This not only fits well to the CNN formulation but it also requires low memory. However, it has a severe limitation that it can not reconstruct convex parts like inside cloth wrinkles. In addition, because the template human model is naked with fingers in hands and feet and deviation mapping can only encode displacement in the normal vector direction, reconstructing shoes or gloves for example is not possible. In the meantime Gabeur \textit{et al.}~\cite{GabeurFMSR2019} proposed a method that predicts depth maps from visible and hidden side by using GAN to "mould" the 3D human body.

We reason that when reconstructing the human body with loose clothes, the implicit volumetric TSDF is the best representation to handle various shapes. We observe that in the regular grid, many of the voxels in the 3D bounding box around the person are actually unnecessary. Therefore, we propose a new tetrahedral 3D shape representation that is able to reduce the memory consumption drastically while allowing to reconstruct high resolution 3D shapes.

\section{Proposed tetrahedral representation}
\label{sec::tetra}
The Truncated Signed Distance Function was first introduced by Curless and Levoy in \cite{curless1996volumetric} to represent 3D surfaces and has been extensively used in modern RGB-D simultaneous localisation and mapping (SLAM) systems \cite{newcombe2015dynamicfusion, newcombe2011kinectfusion, prisacariu2017infinitam, roth2012moving}. 
At any point in the 3D space, the TSDF function takes the signed distance to the 3D surface as value. These values are truncated between $-1$ and $1$ for practical implementation reasons. In general the TSDF is sampled in a regular grid of (rectangular) voxels and the 3D mesh of the surface can be extracted using well established algorithms such as the Marching Cubes \cite{lorensen1987marching} or ray tracing.



\begin{figure}[t]
\centering
\includegraphics[width=1.0\linewidth]{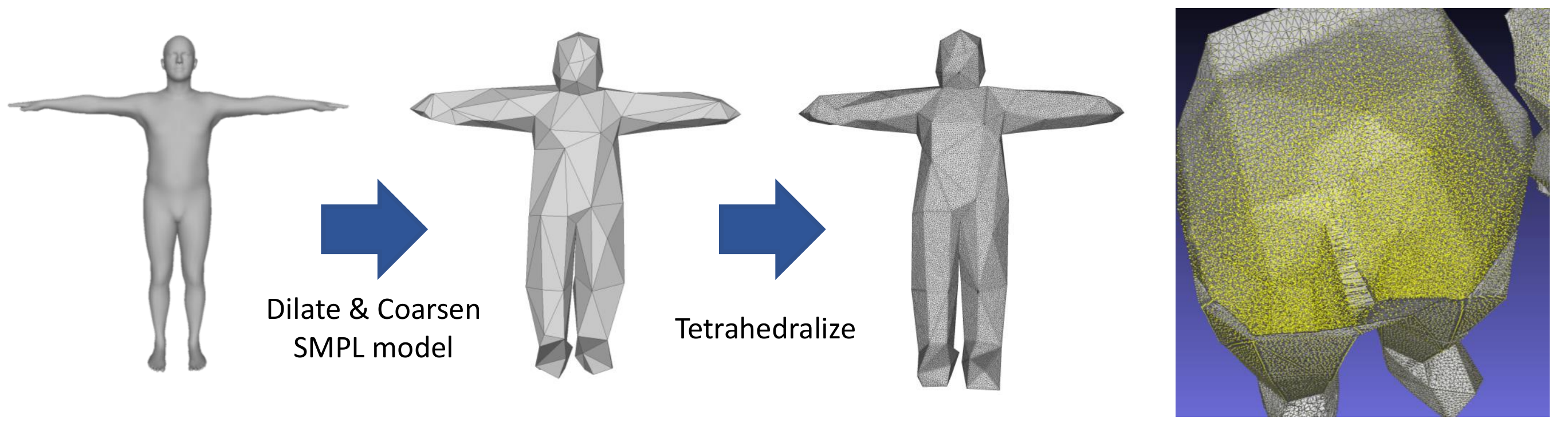}
\caption{Our tetrahedral outer shell for the human body is built from a coarsen version of SMPL template model.}
\label{fig:coarsemodel}
\end{figure}

\begin{figure}[t]
\centering
\includegraphics[width=1.0\linewidth]{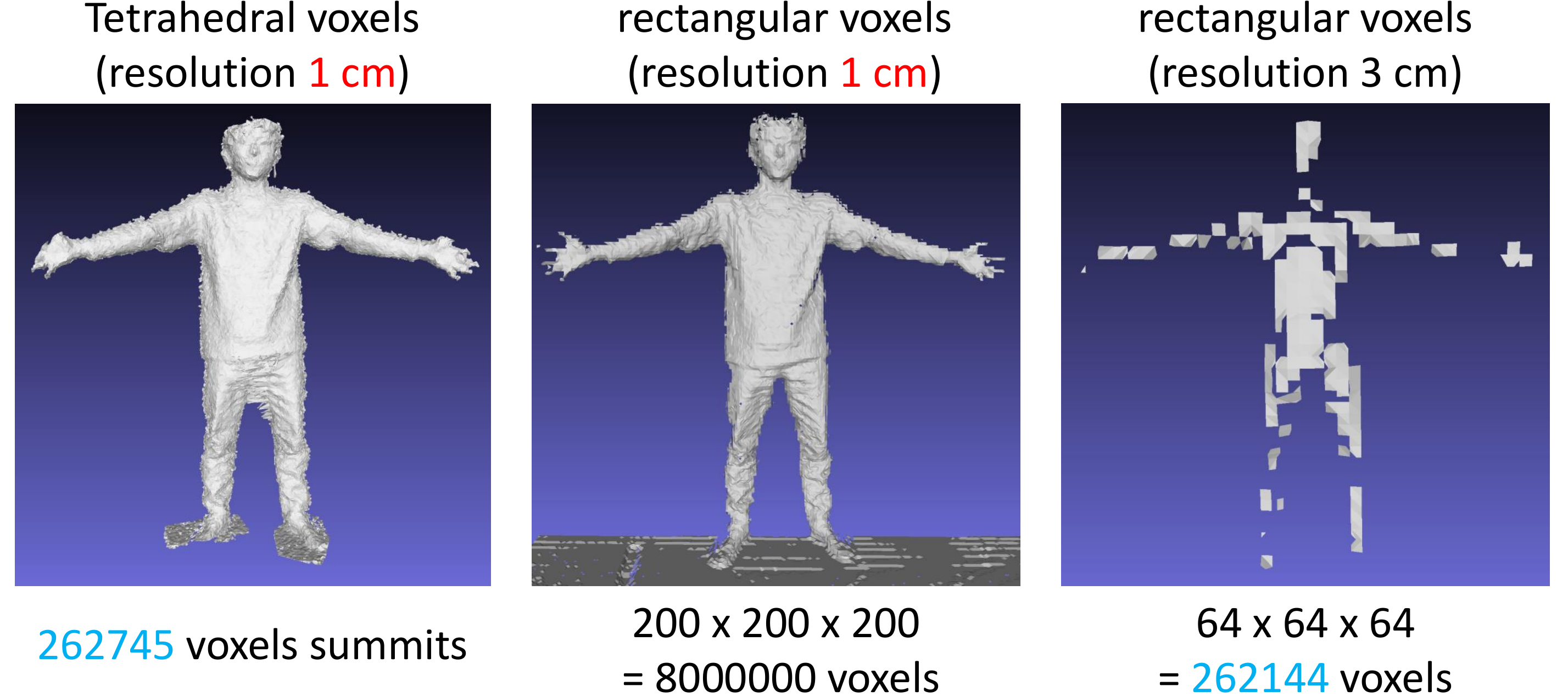}
\caption{Our representation allows us to reconstruct detailed 3D shapes while using a small amount of voxels.}
\label{fig:MCvsMT}
\end{figure}

Our objective here is to reduce the irrelevant field around the human body while covering the meaningful space, which we call the outer shell. We propose to modify the well known SMPL model \cite{SMPL:2015} to create this outer shell. The SMPL model has well defined pose and shape parameters that can be fitted to any 3D human dataset, skeleton or even RGB image (using CNNs for example).

\subsection{Coarse human outer shell}
We propose to inflate the SMPL neutral body model so that once fitted to the input it covers the entire body as well as the loose clothes. We also propose to remove shape details of the SMPL model (such as nose, mouth etc ...). Our pipeline is illustrated in Figure.~\ref{fig:coarsemodel}. We reason that we do not need details at the surface of the outer-shell because the details will be encoded into the TSDF field. Therefore we down-sample the SMPL model, and then up-sample to generate a coarse outer shell with high density vertices\footnote{We used the Blender software (\url{https://www.blender.org}) to do the dilation, down-sampling and up-sampling processes}.


Our proposed outer shell allows us to define a compact 3D space around the person. However, the shape of the outer shell is highly irregular and the standard uniform grid discretization of the 3D space in rectangular voxels is not possible anymore. Thus we propose using tetrahedral discretization of the 3D space instead of the standard rectangular voxels. Figure~\ref{fig:coarsemodel} illustrates the tetrahedral grid created from the outer shell. Note that the outer shell is built only once and then fitted to any person. As a consequence, the same tetrahedral grid can be used for any person.

Figure~\ref{fig:MCvsMT} shows the advantage of our proposed tetrahedral representation over the standard uniform grid for human body 3D representation. Our proposed representation requires significantly less amount of voxels to represent a 3D surface while keeping the same density of points in the extracted surface. Note that the amount of voxels is directly related to the number of parameters in the CNN network. In practice, there are two main differences with the standard rectangular TSDF representation: (1) the TSDF value in the standard volumetric representation is stored at the center of each voxel, whereas, in out case, the value is stored at the summits of each tetrahedra. (2) Naturally, we prefer Marching Tetrahedra algorithm \cite{Shirley:1990:PAD:99307.99322} instead of Marching Cubes to extract the 3D surface.


\begin{figure}[t]
\centering
\includegraphics[width=1.0\linewidth]{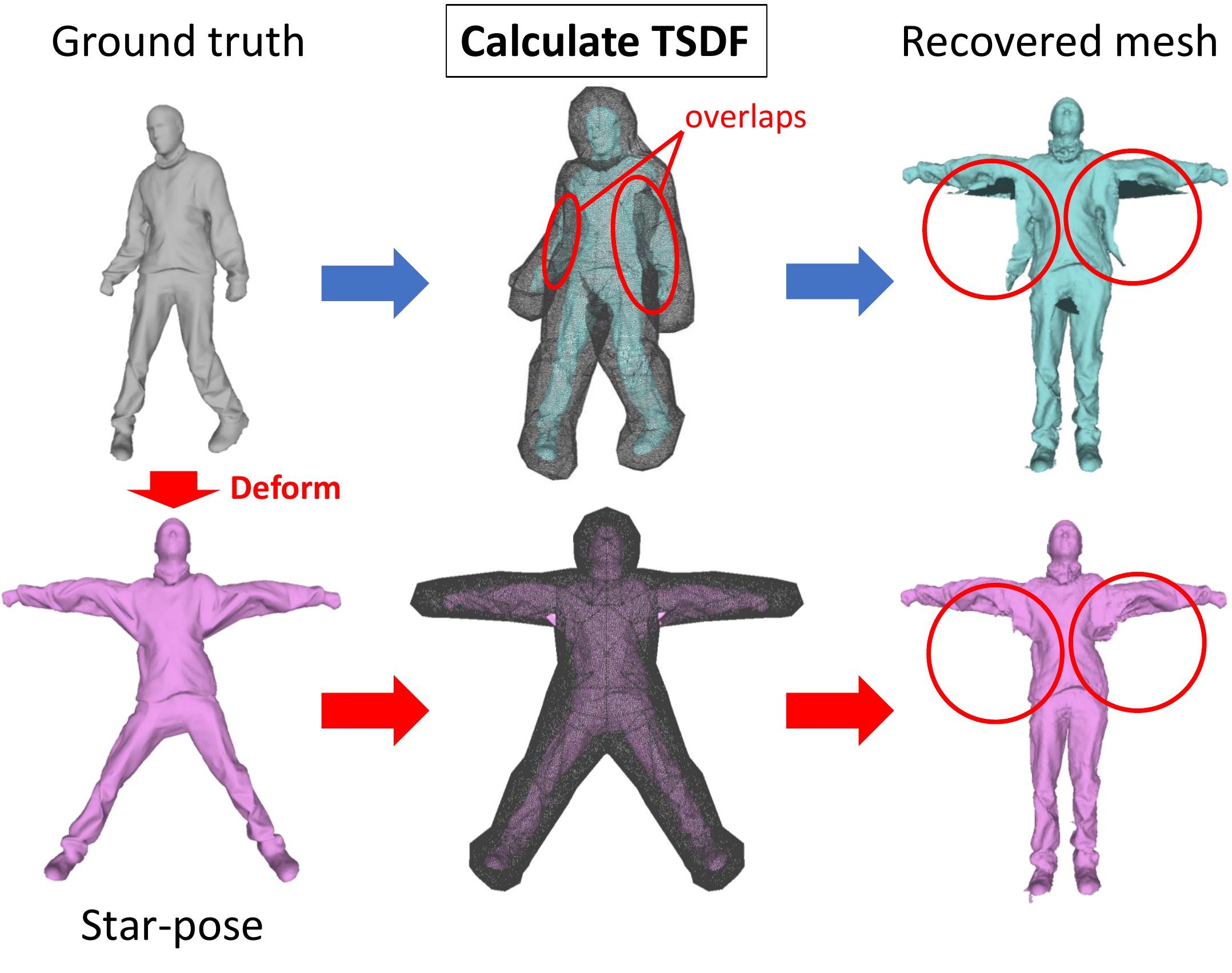}
\caption{After fitting the coarse SMPL model to the 3D scan we calculate the TSDF values from the star-pose warped ground truth 3D mesh.}
\label{fig:fitting}
\end{figure}

\begin{figure*}[t]
\centering
\includegraphics[width=1.0\linewidth]{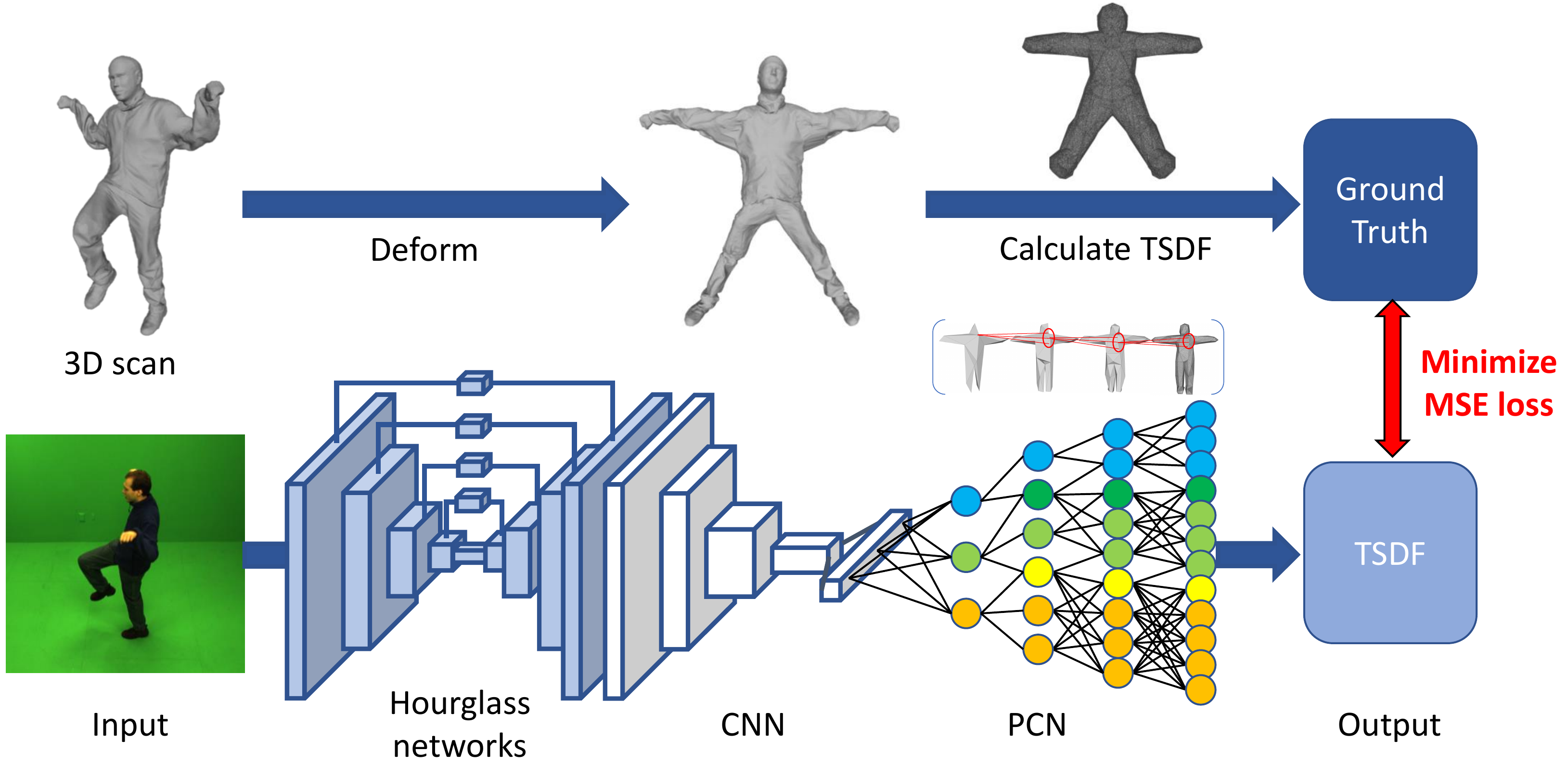}
\caption{Our proposed network is a combination of CNN and PCN. Given a 3D pose data, it allows us to regress the 3D shape of a person wearing clothes in a end-to-end fashion.}
\label{fig:network_training}
\end{figure*}


\subsection{TSDF computation inside tetrahedra}
In order to train a CNN to regress the TSDF values inside our proposed outer shell from a given 2D image, we must first create a ground truth dataset to supervise the learning process. In our case, the training dataset consists of a set of pairs of one 2D image and one corresponding dense tetrahedral TSDF field (\ie, the set of TSDF values for each voxel summit). To build a large amount of dense TSDF fields from a set of publicly available ground truth human 3D scans we need an efficient algorithm. In this section we detail our proposed algorithm to generate such ground truth training dataset.

Firstly, we fit the coarse model to the GT 3D scan by optimizing the SMPL pose and shape parameters given the GT 3D skeleton. Then we compute the TSDF value for each voxel of the outer shell. Here arise two problems: (1) the GT 3D mesh is sparse, so the standard signed distance between a voxel summit and its closest point is significantly different than the signed distance to the surface. (2) In some poses parts of the coarse model may overlap, which may result in ghost effects (for example, a part of the torso may be encoded into the voxels that correspond to the arm).

To solve the first problem we compute the TSDF values as the point-to-plane distance between the voxel summit and the tangent plane of the closest point. This is a reasonable approximation of the point-to-surface signed distance (and it can be computed quickly) when the voxel is close to the surface. However, the approximation may become completely wrong when the voxel is far from the surface. To overcome this, we truncate the TSDF values based on a threshold on the euclidean point-to-point distance between the voxel summit and its closest point.

\begin{equation}
    TSDF(v) = \left \{
        \begin{array}{ll}
             \frac{\hat{n} \cdot (v - \hat{v})}{\tau} & \mbox{if } \|v-\hat{v}\|_2 \leq \tau\\
             \sigma(\hat{n} \cdot (v - \hat{v})) & \mbox{if } \|v-\hat{v}\|_2 > \tau
        \end{array}{}
     \right. ,
\end{equation}{}
where $v$ is a voxel summit, $\hat{v}$ is the closest point of $v$ in the GT 3D mesh, and $\hat{n}$ the normal vector at $\hat{v}$. The threshold $\tau$ is set to $3$ cm in the experiments and the function $\sigma(\cdot)$ returns the sign of its argument.

In order to solve the second problem we compute the TSDF values from the 3D scan warped into a star pose, where the body parts of the outer shell do not penetrate each other (see fig.\ref{fig:fitting}). To this end we first connect each vertex of the GT 3D mesh to its corresponding skeletal nodes with appropriate weight. Then we warp the 3D vertex coordinates to the T-shape by blending the 3D transformation of the attached skeletal joints. Once all vertices have been warped, we compute the dense TSDF field of the outer-shell as explained above.

\section{Detailed 3D shape regression}
\label{sec:network}
We propose an end-to-end regression network to estimate the TSDF values in the tetrahedral volume from a single image. Our proposed network takes as input a single 2D image and outputs the dense TSDF field. From this TSDF field a detailed 3D surface of the human can be extracted by using the Marching Tetrahedra algorithm. The estimated 3D mesh can be re-posed using SMPL pose parameters obtained by any state-of-the-art 3D pose estimation method. 

The standard approach to regress TSDF in volumetric data is to use stacked hourglass networks \cite{Newell2016StackedHN}. In this way, the input image is encoded into a feature vector, which is then decoded into the volumetric grid. To this end, CNNs are used to build the network with the help of the well organized data into uniform grids of pixels and voxels. In our case the volumetric data is embedded into the tetrahedral mesh, which does not have the uniform grid organization. As a consequence, state-of-the-art CNN hourglass architecture cannot be used directly. 

We propose a new hourglass structure to regress the TSDF value of each vertex in the tetrahedral voxel model. Our proposed network combines CNN to encode the input image and a new Part Connection Network (PCN) to decode the feature vector (Figure \ref{fig:network_training} illustrates our proposed network). The originality of our network resides in the later part. We propose to build several tetrahedral layers by down-sampling the full resolution outer shell volumetric model (see fig.~\ref{fig:network_training}), and then we propagate the features in the upper layer to output the TSDF field. Note that the number of voxels in the tetrahedral model is too large to directly use a fully connected network (the number of parameters near the final layer consumes about $90$GB of memory in the case of a fully connected network). Instead we propose a Part Connected Network with partially connected layers, where connections between successive layers are done only in-between the same body parts (which consumes only $0.025$ GB of memory).

For each layer $l$ we define the partial connections using the following sparse adjacency matrices.

\begin{eqnarray}
A_1^l =&
\begin{bmatrix}
A(1)^l\\
A(2)^l\\
\vdots\\
A(n_{\rm out})^l
\end{bmatrix}.
\end{eqnarray}

\begin{eqnarray}
A(n)^l =&
\begin{bmatrix}
a_{11} & a_{12} & \cdots & a_{1n_{\rm in}}\\
a_{21} & a_{22} & \cdots & a_{2n_{\rm in}}\\
\vdots & \vdots & \ddots & \vdots \\
a_{m1} & a_{m2} & \cdots & a_{mn_{\rm in}} 
\end{bmatrix},
\end{eqnarray}
\begin{equation}
{\rm where}\quad a_{ij} = \quad
\left \{
\begin{array}{l}
1 {\rm \ if \ } j=adj(n)[i] \\
0 {\rm \ otherwise}.
\end{array}
\right.\nonumber
\end{equation}

\begin{eqnarray}
A_2^l =&
\begin{bmatrix}
1 & \cdots & 1 & 0 & \cdots & 0 & 0 & \cdots & 0\\
0 & \cdots & 0 & 1 & \cdots & 1 & 0 & \cdots & 0\\
\vdots & & & & \ddots& & & & \vdots\\
0 & \cdots & 0 & 0 & \cdots & 0 & 1 & \cdots & 1
\end{bmatrix},
\end{eqnarray}
where $n_{\rm in}, n_{\rm out}$ is the number of input and output nodes, $m$ is the number of adjacent nodes for the $n$-th output node, and $adj(n)$ is the list of indices of input nodes that are connected to the $n$-th output node (and $adj(n)[i]$ is the $i^{th}$ element in the list). There is no guarantee that all of the output nodes have the same number of adjacent nodes, so we flatten the input features by using the matrix $A_1^l$, and reshape it to the shape of the output features by using $A_2^l$.

With these matrices, the features from two successive layers $(l,l+1)$ are transmitted as follows.
\begin{equation}
f_{l+1}=\sigma(A_2^l (A_1^l f_{l} \circ W_l)) ,
\end{equation}
where $f_{l}, f_{l+1}$ are the input and output features of the layer and $W_l$ is the variable weight matrix for all edges. Also, $ \circ $ denotes the element-wise product of two matrices and $\sigma$ denotes the activation function. 

To define the adjacency matrices $A_1^l$ and $A_2^l$ we need to identify the adjacency lists $L$ that connect the input nodes and the output nodes of all layers. To create these adjacency lists, we focus on the locality of the nodes in the tetrahedral model. Concretely, each node in the $l^{th}$ layer is connected to the k-nearest neighbors in the $l+1^{th}$ layer (we used $k=5$ in our experiments). We reason that the tetrahedral voxel model has a human body shaped graph structure and that we can consider that distant nodes have a weak relationship with each other (like between toes and fingertips). As a consequence, by connecting only the near nodes to the next layer and not connecting the distant nodes, the number of learning parameters is drastically reduced without loosing much information from the features from the previous layer. 

We design the network structure so as not to lose features while reducing the number of variable parameters by connecting only the adjacent nodes. However, when connecting the last CNN layer of the first half of the network to the first PCN layer of the latter half, there is no notion of adjacency because nodes from the CNN layer do not have the shape of the human body. Therefore, we use a fully connected layer between the CNN and the PCN networks. 


\section{Experiments}
\label{sec:experiments}

We evaluate qualitatively and quantitatively the ability of our proposed method to reconstruct 3D shapes of the human body from a single image by using publicly available datasets.

In all our experiments, we used a volumetric resolution (\ie, average distance between adjacent summits in the tetrahedral volume) of about $1$ cm (which corresponds to about $2.6\times 10^5$ voxel summits). We evaluated our network on SURREAL \cite{varol17_surreal} and Articulated \cite{vlasic2008articulated} datasets. In the evaluation using SURREAL we strictly followed the protocol as explained by the authors in \cite{varol17_surreal}. Our network was tested on subjects not appearing in the training dataset. This allows us to test the generalizability of our proposed method. In Articulated, we trained our network on $80\%$ of the data and tested on the remaining $20\%$. Our method was tested on poses (and thus deformations) not appearing in the training dataset. This allows us to test the ability of our proposed methods to reconstruct detailed 3D shapes such as clothes wrinkles. In the training, we employed the mean square error loss function in the last layer and reLU as an activation function in the hidden layers. It took about 3-4 hours to train the network for Articulated dataset, with using a batch size of $5$ and a single GTX 1080 GPU. 

We compared our proposed method with other recent works that have made their code publicly available (\cite{Alldieck_2019_ICCV, SMPL-X:2019, varol18_bodynet}). Note that for Tex2Shape \cite{Alldieck_2019_ICCV} only the code for testing is available and so we built a discriminator network for the training of Tex2Shape referring to their paper. 


\begin{table}[tb]
\centering
  \caption{Quantitative comparison of results obtained with our method, BodyNet and Tex2Shape on the SURREAL \cite{varol17_surreal} (naked) and Articulated \cite{vlasic2008articulated} (clothed) datasets.}
  \label{tab::result}
  \begin{tabular}{|l|c|c|}\hline
    Chamfer (cm) & SURREAL & Articulated\\\hline\hline
    SMPLify-x & n.a. & 9.61\\\hline
    BodyNet & 6.38 & 7.22\\\hline
    Tex2Shape & n.a. & 0.72\\\hline
    Ours & \textbf{5.14} & \textbf{0.43}\\\hline
  \end{tabular}

\end{table}

\subsection{Comparative evaluation on SURREAL dataset}

\begin{figure}[b]
\centering
\includegraphics[width=1.0\linewidth]{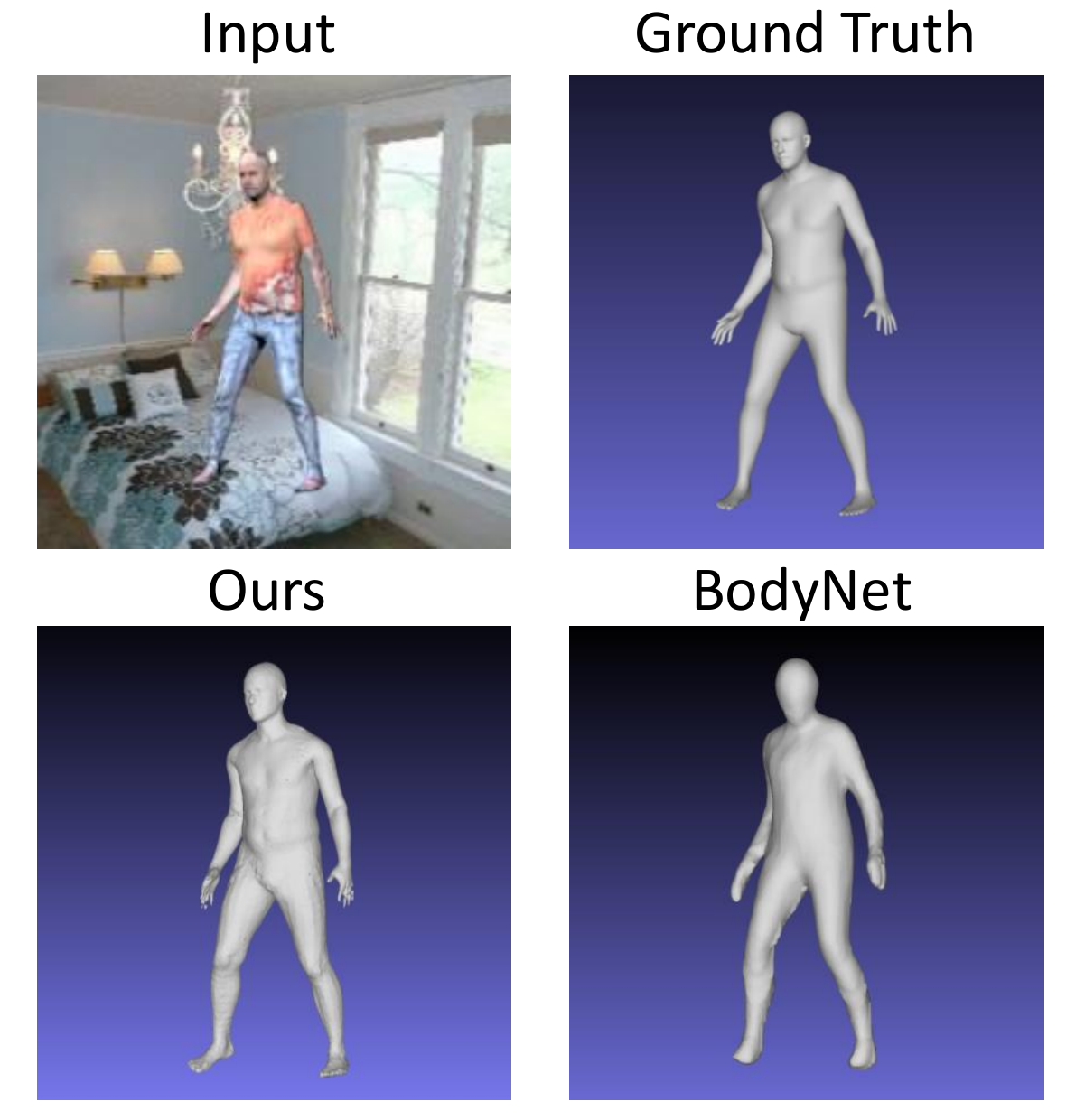}
\caption{An example of a 3D mesh obtained with our method and with BodyNet on the SURREAL dataset.}
\label{fig:surreal}
\end{figure}

\begin{figure*}[t]
\centering
\includegraphics[width=1.0\linewidth]{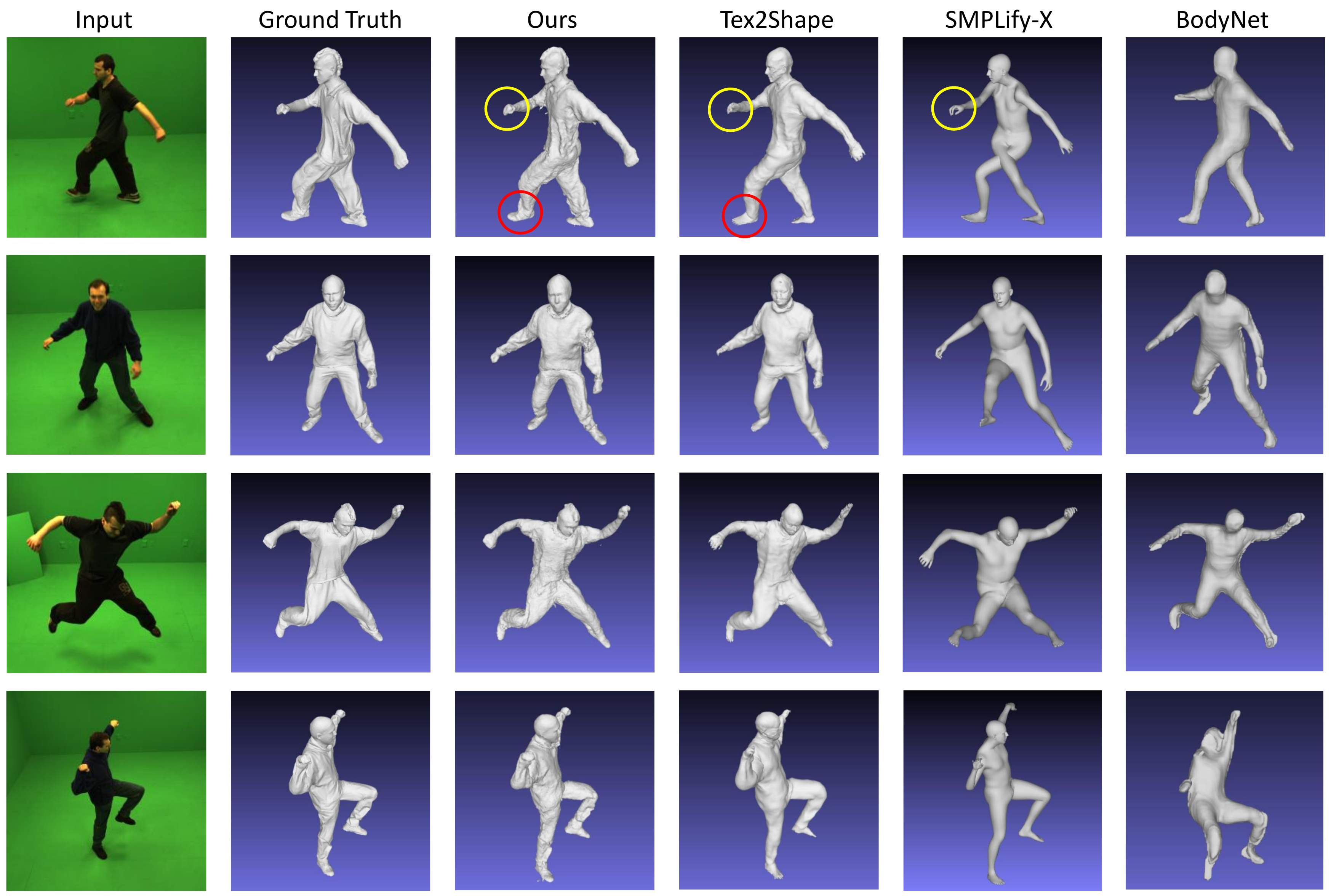}
\caption{Comparative reconstruction results obtained with our proposed method and other related works. From left to right: input image, ground truth, our method, Tex2Shape \cite{Alldieck_2019_ICCV}, SMPLiFy-X \cite{SMPL-X:2019}, BodyNet \cite{varol18_bodynet}}
\label{fig:result1}
\end{figure*}

We compare our proposed method with BodyNet \cite{varol18_bodynet} to confirm the advantage of using our proposed tetrahedral volumetric representation over the classic uniform rectangular grid. We trained and tested our network on the SURREAL dataset \cite{varol17_surreal}, which was used in \cite{varol18_bodynet}. Figure \ref{fig:surreal} shows the qualitative comparative results obtained with our proposed method and BodyNet. As we can see our proposed CNN-PCN network was able to successfully reconstruct the detailed 3D shapes of the human body from only one single image. Note that BodyNet estimates both pose and shape while our method only estimates the 3D shape. We used HMR\cite{hmrKanazawa17} pose estimation results as pose parameters to show our results in the same pose as the input image.

Table \ref{tab::result} shows the quantitative comparison. For the metric we used the Chamfer distance between the ground truth 3D scans and the reconstructed 3D meshes. As we can see from these results our proposed method was able to reconstruct accurate dense 3D shapes of the human body from only one single image. We obtained better results that BodyNet because our proposed tetrahedral representation allows for reconstruction at higher resolution. 

\subsection{Comparative evaluation on Articulated dataset}

\begin{figure*}[h]
\centering
\includegraphics[width=1.0\linewidth]{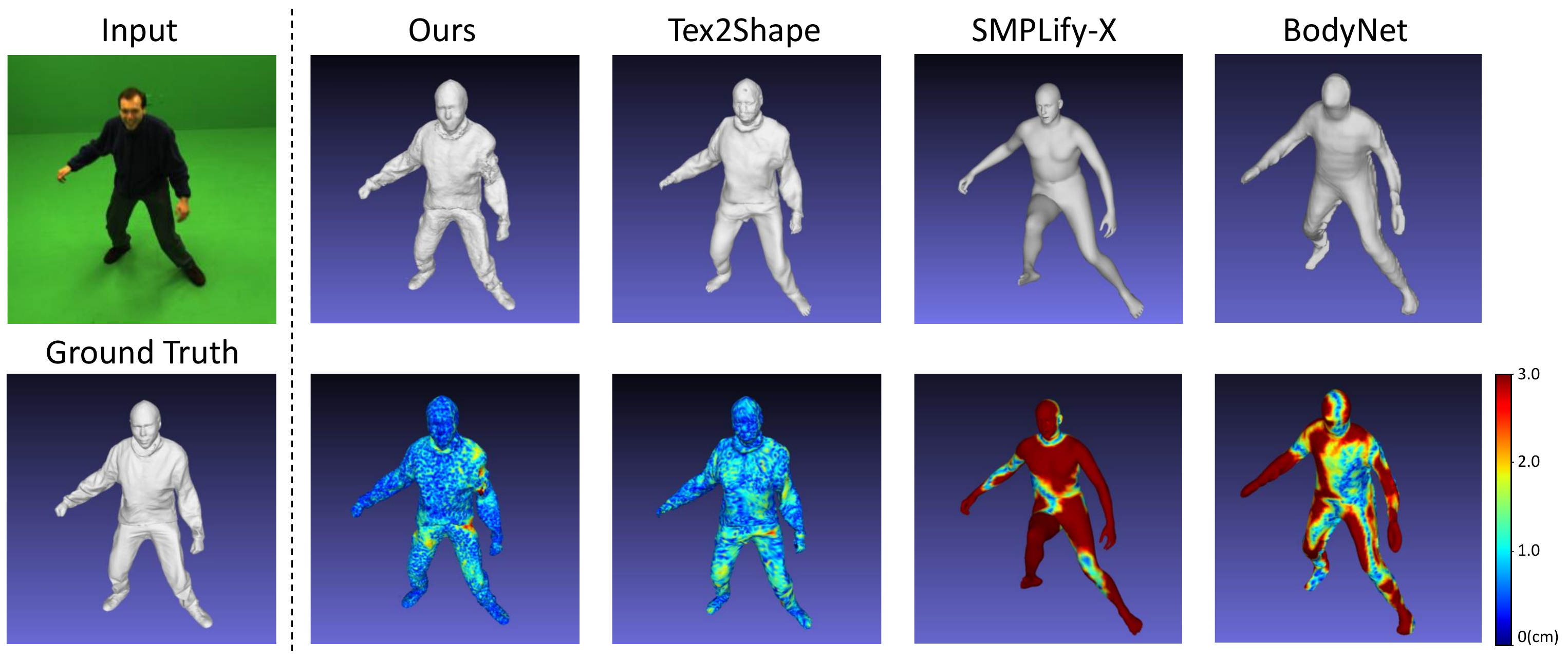}
\caption{Visualization of the Euclidean distances between the vertices of reconstructed 3D mesh and the closest points in the GT scan. These errors are represented with heatmaps and mapped over the reconstructed 3D mesh for each method.}
\label{fig:heatmaps}
\vspace*{-\baselineskip}
\end{figure*}

To confirm the advantages of our proposed method for reconstructing detailed 3D shapes of humans with loose clothes from a single image we compared our method with the most recent state-of-the-art method Tex2Shape \cite{Alldieck_2019_ICCV}. For qualitative and quantitative evaluation, we used the publicly available dataset called Articulated \cite{vlasic2008articulated} that contains sequences of human wearing loose clothes in many poses with ground truth 3D scans. 

Unfortunately, the dataset used in \cite{Alldieck_2019_ICCV} is not publicly available so we could not directly compare our proposed method with \cite{Alldieck_2019_ICCV} on their own dataset. We built a discriminator network for the training of Tex2Shape referring to their paper \cite{Alldieck_2019_ICCV} and trained their network on ARTICULATED dataset using the exact same train/test split as used with our proposed method.
The comparative results shown in Figure \ref{fig:result1} and Table \ref{tab::result} shows advantages of our proposed method compared with Tex2Shape. Note that as in \cite{Alldieck_2019_ICCV} we show the results re-posed with the ground truth 3D pose. However, any third party 3D pose estimation could also be used to estimate the 3D pose from the input image (\cite{moreno20173d, wang2018robust}). 
Figure \ref{fig:result1} shows some representative results from the test dataset.

We also compared the results obtained with our method with those obtained with other closely related works. As seen in figures \ref{fig:result1} and \ref{fig:heatmaps}, our proposed method outperforms all other previous works. Figure \ref{fig:heatmaps} shows the heat maps of errors of the 3D reconstructed models obtained by our method and other related works. As we can see our proposed network could successfully recover the details of the loose clothes, even in occluded areas. As we can see in Table \ref{tab::result} we observed an average error around $0.5$ cm in the 3D meshes reconstructed with our proposed method. 

As we can see in the circled areas of figure \ref{fig:result1}, the strongest and clearer advantage of our proposed method over the related works is that our proposed method can retrieve both hand details such as fingers in the correct position (no details can be obtained with BodyNet) or in the inverse reconstruct the shoes around the feet (fingers still remain with Tex2Shape, giving an unpleasant effect for people wearing shoes). Moreover, with our proposed method we do not need to estimate finger poses. As we can see in Figure \ref{fig:result1} although the hand pose is unknown (this is why the hand is always open in the results obtained with Tex2Shape) our proposed method could successfully reconstruct the detailed hand in the correct pose.

\subsection{Network analysis}
\begin{figure}[t]
\includegraphics[width=1.0\linewidth]{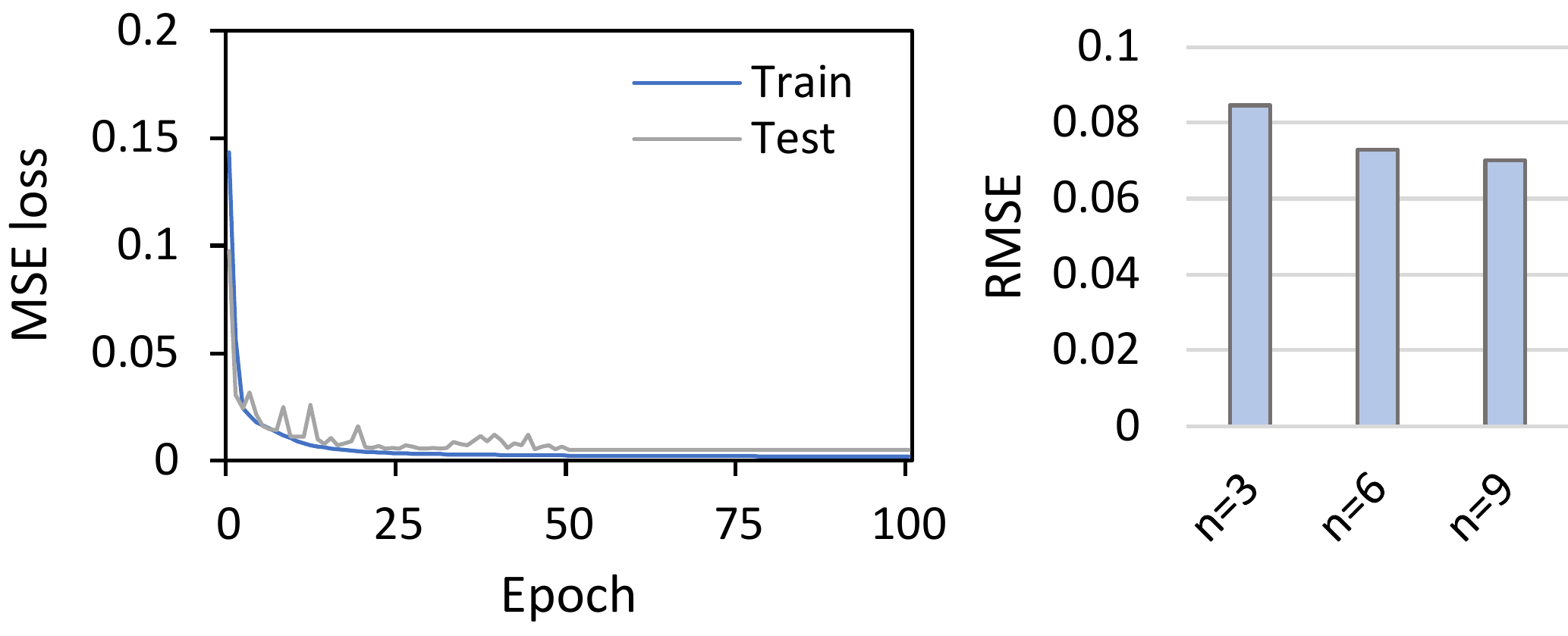}
\caption{Left: learning curve of our network on ARTICULATED dataset. Right: we reduced the connection of the nodes from original (n=9) to n=3, n=6. 
}
\label{fig:graphs}
\vspace*{-\baselineskip}
\end{figure}

We analysed the performance of our network by changing the parameters. Figure \ref{fig:graphs} (a) shows the convergence speed of our network during training. Figure \ref{fig:graphs} (b) shows the performance of our network when changing the number of connections in the PCN. As expected, there is a trade-off between the number of connections (\ie, memory consumption) and accuracy.

\section{Conclusion}
\label{sec:conclusion}
We proposed a method for reconstructing fine detailed 3D shapes of human body wearing loose clothes from a single image. We introduced a novel 3D human body representation based on a tetrahedral TSDF field embedded into a coarse human outer shell. We also designed CNN-PCN network to regress the tetrahedral TSDF field in an end-to-end fashion. Our qualitative and quantitative comparative experiments using public datasets confirmed the ability of our proposed method to reconstruct detailed shapes with loose clothes. The results demonstrated out-performance of our method against the current state-of-the-art on these datasets. Several possible improvements are left for future work, such as combining our PCN with the Graph Convolutional Network \cite{Defferrard:2016:CNN:3157382.3157527}\cite{Yan2018SpatialTG}.

\section*{Acknowledgement}\vspace*{-6pt}

A part of this work was supported by JSPS KAKENHI Grant Number JP19K20297 and Microsoft CORE 16 Grant.

{\small
\bibliographystyle{ieee_fullname}
\bibliography{main}

\begin{thebibliography}{10}\itemsep=-1pt

\bibitem{Alldieck_2019_ICCV}
Thiemo Alldieck, Gerard Pons-Moll, Christian Theobalt, and Marcus Magnor.
\newblock Tex2shape: Detailed full human body geometry from a single image.
\newblock In {\em The IEEE International Conference on Computer Vision (ICCV)},
  October 2019.

\bibitem{bualan2008naked}
Alexandru~O B{\u{a}}lan and Michael~J Black.
\newblock The naked truth: Estimating body shape under clothing.
\newblock In {\em European Conference on Computer Vision}, pages 15--29.
  Springer, 2008.

\bibitem{balan2007detailed}
Alexandru~O Balan, Leonid Sigal, Michael~J Black, James~E Davis, and Horst~W
  Haussecker.
\newblock Detailed human shape and pose from images.
\newblock In {\em 2007 IEEE Conference on Computer Vision and Pattern
  Recognition}, pages 1--8. IEEE, 2007.

\bibitem{Bogo:ECCV:2016}
Federica Bogo, Angjoo Kanazawa, Christoph Lassner, Peter Gehler, Javier Romero,
  and Michael~J. Black.
\newblock Keep it {SMPL}: Automatic estimation of {3D} human pose and shape
  from a single image.
\newblock In {\em Computer Vision -- ECCV 2016}, Lecture Notes in Computer
  Science. Springer International Publishing, Oct. 2016.

\bibitem{cashman2012shape}
Thomas~J Cashman and Andrew~W Fitzgibbon.
\newblock What shape are dolphins? building 3d morphable models from 2d images.
\newblock {\em IEEE transactions on pattern analysis and machine intelligence},
  35(1):232--244, 2012.

\bibitem{choy20163d}
Christopher~B Choy, Danfei Xu, JunYoung Gwak, Kevin Chen, and Silvio Savarese.
\newblock 3d-r2n2: A unified approach for single and multi-view 3d object
  reconstruction.
\newblock In {\em European conference on computer vision}, pages 628--644.
  Springer, 2016.

\bibitem{curless1996volumetric}
Brian Curless and Marc Levoy.
\newblock A volumetric method for building complex models from range images.
\newblock 1996.

\bibitem{de2008performance}
Edilson De~Aguiar, Carsten Stoll, Christian Theobalt, Naveed Ahmed, Hans-Peter
  Seidel, and Sebastian Thrun.
\newblock {\em Performance capture from sparse multi-view video}, volume~27.
\newblock ACM, 2008.

\bibitem{Defferrard:2016:CNN:3157382.3157527}
Micha\"{e}l Defferrard, Xavier Bresson, and Pierre Vandergheynst.
\newblock Convolutional neural networks on graphs with fast localized spectral
  filtering.
\newblock In {\em Proceedings of the 30th International Conference on Neural
  Information Processing Systems}, NIPS, pages 3844--3852, USA, 2016. Curran
  Associates Inc.

\bibitem{GabeurFMSR2019}
Valentin Gabeur, Jean-Sébastien~Franco Franco, Xavier Martin, Cordelia Schmid,
  and Gregory Rogez.
\newblock Moulding humans: Non-parametric 3d human shape estimation from single
  images.
\newblock In {\em IEEE/CVF International Conference on Computer Vision (ICCV)},
  2019.

\bibitem{girdhar2016learning}
Rohit Girdhar, David~F Fouhey, Mikel Rodriguez, and Abhinav Gupta.
\newblock Learning a predictable and generative vector representation for
  objects.
\newblock In {\em European Conference on Computer Vision}, pages 484--499.
  Springer, 2016.

\bibitem{guan2009estimating}
Peng Guan, Alexander Weiss, Alexandru~O Balan, and Michael~J Black.
\newblock Estimating human shape and pose from a single image.
\newblock In {\em 2009 IEEE 12th International Conference on Computer Vision},
  pages 1381--1388. IEEE, 2009.

\bibitem{habermann2019livecap}
Marc Habermann, Weipeng Xu, Michael Zollhoefer, Gerard Pons-Moll, and Christian
  Theobalt.
\newblock Livecap: Real-time human performance capture from monocular video.
\newblock {\em ACM Transactions on Graphics (TOG)}, 38(2):14, 2019.

\bibitem{iskakov2019learnable}
Karim Iskakov, Egor Burkov, Victor Lempitsky, and Yury Malkov.
\newblock Learnable triangulation of human pose.
\newblock In {\em International Conference on Computer Vision (ICCV)}, 2019.

\bibitem{Jackson2017LargeP3}
Aaron~S. Jackson, Adrian Bulat, Vasileios Argyriou, and Georgios Tzimiropoulos.
\newblock Large pose 3d face reconstruction from a single image via direct
  volumetric cnn regression.
\newblock {\em 2017 IEEE International Conference on Computer Vision (ICCV)},
  pages 1031--1039, 2017.

\bibitem{hmrKanazawa17}
Angjoo Kanazawa, Michael~J. Black, David~W. Jacobs, and Jitendra Malik.
\newblock End-to-end recovery of human shape and pose.
\newblock In {\em Computer Vision and Pattern Regognition (CVPR)}, 2018.

\bibitem{kolotouros2019learning}
Nikos Kolotouros, Georgios Pavlakos, Michael~J Black, and Kostas Daniilidis.
\newblock Learning to reconstruct 3d human pose and shape via model-fitting in
  the loop.
\newblock In {\em Proceedings of the IEEE International Conference on Computer
  Vision}, pages 2252--2261, 2019.

\bibitem{kraevoy2009modeling}
Vladislav Kraevoy, Alla Sheffer, and Michiel van~de Panne.
\newblock Modeling from contour drawings.
\newblock In {\em Proceedings of the 6th Eurographics Symposium on Sketch-Based
  interfaces and Modeling}, pages 37--44. ACM, 2009.

\bibitem{SMPL:2015}
Matthew Loper, Naureen Mahmood, Javier Romero, Gerard Pons-Moll, and Michael~J.
  Black.
\newblock {SMPL}: A skinned multi-person linear model.
\newblock {\em ACM Trans. Graphics (Proc. SIGGRAPH Asia)}, 34(6):248:1--248:16,
  Oct. 2015.

\bibitem{lorensen1987marching}
William~E Lorensen and Harvey~E Cline.
\newblock Marching cubes: A high resolution 3d surface construction algorithm.
\newblock In {\em ACM siggraph computer graphics}, volume~21, pages 163--169.
  ACM, 1987.

\bibitem{lu2008deformation}
Xiaoguang Lu and Anil Jain.
\newblock Deformation modeling for robust 3d face matching.
\newblock {\em IEEE Transactions on Pattern Analysis and Machine Intelligence},
  30(8):1346--1357, 2008.

\bibitem{moreno20173d}
Francesc Moreno-Noguer.
\newblock 3d human pose estimation from a single image via distance matrix
  regression.
\newblock In {\em Proceedings of the IEEE Conference on Computer Vision and
  Pattern Recognition}, pages 2823--2832, 2017.

\bibitem{newcombe2015dynamicfusion}
Richard~A Newcombe, Dieter Fox, and Steven~M Seitz.
\newblock Dynamicfusion: Reconstruction and tracking of non-rigid scenes in
  real-time.
\newblock In {\em Proceedings of the IEEE conference on computer vision and
  pattern recognition}, pages 343--352, 2015.

\bibitem{newcombe2011kinectfusion}
Richard~A Newcombe, Shahram Izadi, Otmar Hilliges, David Molyneaux, David Kim,
  Andrew~J Davison, Pushmeet Kohli, Jamie Shotton, Steve Hodges, and Andrew~W
  Fitzgibbon.
\newblock Kinectfusion: Real-time dense surface mapping and tracking.
\newblock In {\em ISMAR}, volume~11, pages 127--136, 2011.

\bibitem{Newell2016StackedHN}
Alejandro Newell, Kaiyu Yang, and Jia Deng.
\newblock Stacked hourglass networks for human pose estimation.
\newblock In {\em ECCV}, 2016.

\bibitem{SMPL-X:2019}
Georgios Pavlakos, Vasileios Choutas, Nima Ghorbani, Timo Bolkart, Ahmed A.~A.
  Osman, Dimitrios Tzionas, and Michael~J. Black.
\newblock Expressive body capture: 3d hands, face, and body from a single
  image.
\newblock In {\em Proceedings IEEE Conf. on Computer Vision and Pattern
  Recognition (CVPR)}, 2019.

\bibitem{prisacariu2017infinitam}
Victor~Adrian Prisacariu, Olaf K{\"a}hler, Stuart Golodetz, Michael Sapienza,
  Tommaso Cavallari, Philip~HS Torr, and David~W Murray.
\newblock Infinitam v3: a framework for large-scale 3d reconstruction with loop
  closure.
\newblock {\em arXiv preprint arXiv:1708.00783}, 2017.

\bibitem{riegler2017octnet}
Gernot Riegler, Ali Osman~Ulusoy, and Andreas Geiger.
\newblock Octnet: Learning deep 3d representations at high resolutions.
\newblock In {\em Proceedings of the IEEE Conference on Computer Vision and
  Pattern Recognition}, pages 3577--3586, 2017.

\bibitem{roth2012moving}
Henry Roth and Marsette Vona.
\newblock Moving volume kinectfusion.
\newblock In {\em BMVC}, volume~20, pages 1--11, 2012.

\bibitem{Saito_2019_ICCV}
Shunsuke Saito, Zeng Huang, Ryota Natsume, Shigeo Morishima, Angjoo Kanazawa,
  and Hao Li.
\newblock Pifu: Pixel-aligned implicit function for high-resolution clothed
  human digitization.
\newblock In {\em The IEEE International Conference on Computer Vision (ICCV)},
  October 2019.

\bibitem{Shirley:1990:PAD:99307.99322}
Peter Shirley and Allan Tuchman.
\newblock A polygonal approximation to direct scalar volume rendering.
\newblock In {\em Proceedings of the 1990 Workshop on Volume Visualization},
  VVS '90, pages 63--70, New York, NY, USA, 1990. ACM.

\bibitem{tan2010image}
Guang-hua Tan, Wei Chen, and Li-gang Liu.
\newblock Image driven shape deformation using styles.
\newblock {\em Journal of Zhejiang University SCIENCE C}, 11(1):27, 2010.

\bibitem{tatarchenko2016multi}
Maxim Tatarchenko, Alexey Dosovitskiy, and Thomas Brox.
\newblock Multi-view 3d models from single images with a convolutional network.
\newblock In {\em European Conference on Computer Vision}, pages 322--337.
  Springer, 2016.

\bibitem{tatarchenko2017octree}
Maxim Tatarchenko, Alexey Dosovitskiy, and Thomas Brox.
\newblock Octree generating networks: Efficient convolutional architectures for
  high-resolution 3d outputs.
\newblock In {\em Proceedings of the IEEE International Conference on Computer
  Vision}, pages 2088--2096, 2017.

\bibitem{tulsiani2017multi}
Shubham Tulsiani, Tinghui Zhou, Alexei~A Efros, and Jitendra Malik.
\newblock Multi-view supervision for single-view reconstruction via
  differentiable ray consistency.
\newblock In {\em Proceedings of the IEEE conference on computer vision and
  pattern recognition}, pages 2626--2634, 2017.

\bibitem{varol18_bodynet}
G{\"u}l Varol, Duygu Ceylan, Bryan Russell, Jimei Yang, Ersin Yumer, Ivan
  Laptev, and Cordelia Schmid.
\newblock {BodyNet}: Volumetric inference of {3D} human body shapes.
\newblock In {\em ECCV}, 2018.

\bibitem{varol17_surreal}
G{\"u}l Varol, Javier Romero, Xavier Martin, Naureen Mahmood, Michael~J. Black,
  Ivan Laptev, and Cordelia Schmid.
\newblock Learning from synthetic humans.
\newblock In {\em CVPR}, 2017.

\bibitem{vlasic2008articulated}
Daniel Vlasic, Ilya Baran, Wojciech Matusik, and Jovan Popovi{\'c}.
\newblock Articulated mesh animation from multi-view silhouettes.
\newblock In {\em ACM Transactions on Graphics (TOG)}, volume~27, page~97. ACM,
  2008.

\bibitem{wang2018robust}
Chunyu Wang, Yizhou Wang, Zhouchen Lin, and Alan~L Yuille.
\newblock Robust 3d human pose estimation from single images or video
  sequences.
\newblock {\em IEEE transactions on pattern analysis and machine intelligence},
  41:1227--1241, 2018.

\bibitem{wu2017marrnet}
Jiajun Wu, Yifan Wang, Tianfan Xue, Xingyuan Sun, Bill Freeman, and Josh
  Tenenbaum.
\newblock Marrnet: 3d shape reconstruction via 2.5 d sketches.
\newblock In {\em Advances in neural information processing systems}, pages
  540--550, 2017.

\bibitem{wu2016learning}
Jiajun Wu, Chengkai Zhang, Tianfan Xue, Bill Freeman, and Josh Tenenbaum.
\newblock Learning a probabilistic latent space of object shapes via 3d
  generative-adversarial modeling.
\newblock In {\em Advances in neural information processing systems}, pages
  82--90, 2016.

\bibitem{wu20153d}
Zhirong Wu, Shuran Song, Aditya Khosla, Fisher Yu, Linguang Zhang, Xiaoou Tang,
  and Jianxiong Xiao.
\newblock 3d shapenets: A deep representation for volumetric shapes.
\newblock In {\em Proceedings of the IEEE conference on computer vision and
  pattern recognition}, pages 1912--1920, 2015.

\bibitem{Yan2018SpatialTG}
Sijie Yan, Yuanjun Xiong, and Dahua Lin.
\newblock Spatial temporal graph convolutional networks for skeleton-based
  action recognition.
\newblock In {\em AAAI}, 2018.

\bibitem{yan2016perspective}
Xinchen Yan, Jimei Yang, Ersin Yumer, Yijie Guo, and Honglak Lee.
\newblock Perspective transformer nets: Learning single-view 3d object
  reconstruction without 3d supervision.
\newblock In {\em Advances in Neural Information Processing Systems}, pages
  1696--1704, 2016.

\bibitem{yu2018doublefusion}
Tao Yu, Zerong Zheng, Kaiwen Guo, Jianhui Zhao, Qionghai Dai, Hao Li, Gerard
  Pons-Moll, and Yebin Liu.
\newblock Doublefusion: Real-time capture of human performances with inner body
  shapes from a single depth sensor.
\newblock In {\em Proceedings of the IEEE Conference on Computer Vision and
  Pattern Recognition}, pages 7287--7296, 2018.

\bibitem{zhu2017rethinking}
Rui Zhu, Hamed Kiani~Galoogahi, Chaoyang Wang, and Simon Lucey.
\newblock Rethinking reprojection: Closing the loop for pose-aware shape
  reconstruction from a single image.
\newblock In {\em Proceedings of the IEEE International Conference on Computer
  Vision}, pages 57--65, 2017.

\end{thebibliography}
}

\end{document}